\documentclass[11pt]{article}

\usepackage[final]{acl}

\usepackage{times}
\usepackage{latexsym}

\usepackage[T1]{fontenc}

\usepackage[utf8]{inputenc}

\usepackage{microtype}

\usepackage{inconsolata}

\usepackage{graphicx}
\usepackage{algorithm}
\usepackage{algorithmic}
\usepackage{amsmath}
\usepackage{booktabs}
\usepackage{multicol}
\usepackage{multirow}
\usepackage[table]{xcolor}
\usepackage{listings}
\usepackage[most]{tcolorbox}
\usepackage[dvipsnames]{xcolor}
\usepackage{enumitem}

\title{Coder as Editor: Code-driven Interpretable Molecular Editing}

\newcommand{\abbr}{MECo}
\definecolor{mygrey}{RGB}{200,200,200}

\author{
 \textbf{Wenyu Zhu\textsuperscript{1,3}},
 \textbf{Chengzhu Li\textsuperscript{4}},
 \textbf{Xiaohe Tian\textsuperscript{5}},
 \textbf{Yifan Wang\textsuperscript{6}},
\\
 \textbf{Yinjun Jia\textsuperscript{1}},
 \textbf{Jianhui Wang\textsuperscript{1}},
 \textbf{Bowen Gao\textsuperscript{1,3,7}},
 \textbf{Haichuan Tan\textsuperscript{1,3,7}},
\\
 \textbf{Ya-Qin Zhang\textsuperscript{1}},
 \textbf{Wei-Ying Ma\textsuperscript{1}},
 \textbf{Yanyan Lan\textsuperscript{1,2,3,*}}
\\
 \footnotesize
 \begin{tabular}{c}
 \textsuperscript{1}Institute for AI Industry Research,
 Tsinghua University, Beijing, China \\
 \textsuperscript{2}Beijing Frontier Research Center for Biological Structure,
 Tsinghua University, Beijing, China \\
 \textsuperscript{3}AIPher Technology Co., Ltd., Beijing, China \\
 \textsuperscript{4}Zhili College, Tsinghua University, Beijing, China \\
 \textsuperscript{5}School of Pharmaceutical Sciences, Peking University, Beijing, China \\
 \textsuperscript{6}College of AI,
 Tsinghua University, Beijing, China \\
 \textsuperscript{7}Department of Computer Science and Technology, Tsinghua University, Beijing, China \\
 \textsuperscript{*}\textbf{Correspondence:}
 \href{mailto:lanyanyan@air.tsinghua.edu.cn}{lanyanyan@air.tsinghua.edu.cn}
 \end{tabular}
}

\begin{document}
\maketitle
\begin{abstract}
Molecular design is a central task in drug discovery that requires precise structural reasoning and domain knowledge. While large language models (LLMs) have shown promise in generating high-level editing intentions in natural language, they often struggle to faithfully execute these modifications, particularly when operating on non-intuitive representations like SMILES. We introduce \abbr{}, a framework that bridges reasoning and execution by translating editing actions into executable code. \abbr{} reformulates molecular modification for LLMs as a cascaded framework: generating human-interpretable editing intentions from a molecule and design objective, followed by translating those intentions into executable structural edits via code generation. Our approach achieves over 98\% accuracy in reproducing held-out realistic edits derived from chemical reactions and target-specific compound pairs. On downstream benchmarks spanning physicochemical properties and target activities, \abbr{} substantially improves consistency by 38-86 percentage points to 90\%+ and achieves higher success rates over SMILES-based baselines while preserving structural similarity. By aligning intention with execution, \abbr{} enables consistent, controllable and interpretable molecular design, laying the foundation for high-fidelity feedback loops and collaborative human–AI workflows in drug discovery.
\end{abstract}

\section{Introduction}


\begin{figure*}[ht]
\begin{center}
\includegraphics[width=0.95\textwidth]{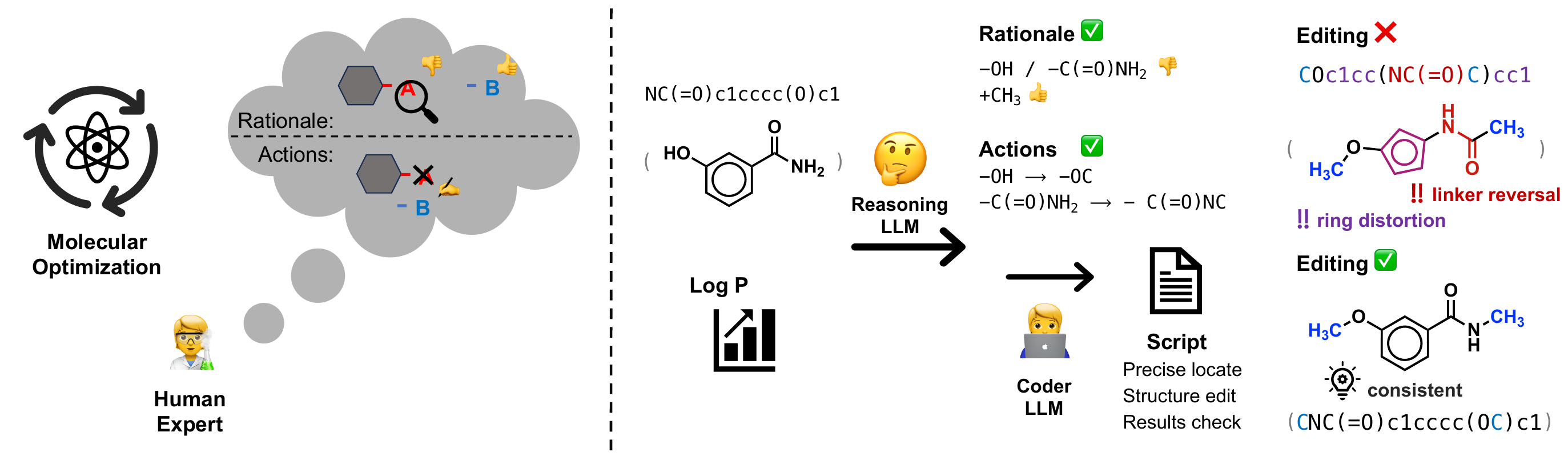}
\end{center}
\caption{Motivation for \abbr{}: bridging reasoning and execution in molecular optimization through code generation. 
Left: Human chemists reason over molecular graphs to design targeted edits, such as modifying a functional group to adjust polarity or introducing new interactions, and annotate them directly on the structure. 
Top right: Reasoning LLMs can generate similar high-level rationales and editing actions, but often struggle to execute them faithfully due to the limitations of sequential molecular representations like SMILES. 
Bottom right: \abbr{} addresses this gap by introducing a coder LLM that translates editing actions into executable code, enabling precise, interpretable, and reproducible structural edits on the molecular graph.}
\label{fig:graph_abs}
\end{figure*}

Recent reasoning-centric large language models (LLMs) have demonstrated notable progress in scientific problem-solving, particularly in programming~\citep{wang2023codet5+}, mathematics~\citep{lewkowycz2022solving}, and chemistry~\citep{zhang2024chemllm}. 
In chemistry, their success on question answering tasks highlights an ability to encode domain knowledge~\citep{hou_benchmarking-gpt5-biomed_2025}, but applying this knowledge to tasks like molecular modification, which require both chemical reasoning and precise structural control, remains a challenge.

A growing number of studies have applied LLMs to molecular modification by prompting them to generate optimized molecules in the form of SMILES strings, conditioned on an input structure and a desired property goal (e.g., increased solubility or binding affinity)~\citep{liu2024conversational, ye2025drugassist}. While intuitive, this generation paradigm faces major limitations. First, generated SMILES may not always be chemically plausible. Second, even valid outputs may diverge from the design rationale articulated by reasoning LLMs, undermining interpretability and expert trust. Such misalignments hinder human-in-the-loop workflows where reproducibility and verifiability are critical, and break the feedback loop required for iterative model refinement. Moreover, extensive uncontrolled edits may severely compromise synthetic feasibility by disrupting established synthetic routes and limiting the reuse of existing intermediates, thereby increasing the practical cost and difficulty of experimental validation.

At the root of these limitations is a modality mismatch: SMILES is a linearized encoding of molecular graphs, originally designed for compact storage in cheminformatics systems. A single molecule may correspond to many valid SMILES depending on the traversal path, and even small structural edits can cause large, unintuitive changes in the string. In contrast, chemists typically reason over molecular graphs and use graphical tools instead of directly modifying SMILES, making precise control difficult for LLMs.

To bridge this gap, we propose using code as an intermediate representation, a domain where LLMs have already demonstrated strong proficiency. Rather than generating molecular structures directly as SMILES, we reformulate molecular editing as a code generation task: the LLM produces executable scripts (e.g., using RDKit~\citep{landrum_rdkit_2013}) that specify verifiable structural modifications. This approach leverages the strengths of LLMs and enables faithful execution of editing intentions to improve controllability, reproducibility, and transparency in molecular design.


We introduce \textbf{\abbr} (\textbf{M}olecular \textbf{E}diting via \textbf{Co}de generation), a framework that translates high-level design rationales into structured editing programs. We fine-tune a compact code-oriented language model solely on synthetic data generated by limited moiety substitutions on random molecules. The model achieves 98\% accuracy on realistic edits derived from reaction and bioactivity datasets, significantly outperforming a same-size general-purpose language model fine-tuned to directly generate SMILES. When using a reasoning LLM such as DeepSeek-R1~\citep{guo_deepseek-r1_2025} as the upstream component, \abbr{} substantially improves the consistency between editing intentions and resulting structures by 38–86 percentage points to 90\%+, leading to higher success rates while preserving structural similarity across multiple property and activity optimization benchmarks.

By bridging natural language reasoning and structural molecular modification through code, \abbr{} enables consistent, controllable and interpretable molecular modification, bringing LLM-guided molecule design closer to real-world scientific application. Our main contributions are:
\begin{itemize}[noitemsep, nolistsep]
\item \textbf{Code-based formulation for molecular modification.} We introduce \textsc{MECo}, a novel framework that recasts molecular editing as a code generation task, enabling LLMs to translate natural language intentions into precise, verifiable, and executable modifications.
\item \textbf{Scalable data construction for training and evaluation.} We develop a scalable pipeline for constructing both synthetic and realistic editing samples, combining programmatic moiety replacement with edit extraction from chemical reactions and bioactive molecules.
\item \textbf{Generalization to realistic molecular transformations.} We show that a code LLM trained solely on synthetic edits generalizes effectively to real-world modifications, achieving over 98\% accuracy on both reaction- and activity-derived edits.
\item \textbf{Superior molecule optimization performance.} \textsc{MECo} outperforms direct generation baselines across property and activity benchmarks, with notably \emph{double the structure–intention consistency}, enhancing interpretability and reliability.
\end{itemize}

\section{Related work}

Despite the widespread adoption of SMILES for sequence-based molecular representation~\citep{weininger1988smiles,gomez2018automatic,segler2018generating,schwaller2019molecular}, it poorly aligns with natural language because small structural modifications can cause disproportionately large string differences, and it suffers from syntactic non-uniqueness for a single molecule~\citep{merz2020generative}. 
Conversely, while graph-based optimization approaches~\citep{jin2018junction,shi2020graphaf,jin_multi-objective_2020,vignac_digress_2022} bypass string syntax constraints, they fundamentally rely on oracle-driven feedback~\citep{gao2022sample} to explore the property landscape.

LLMs have been investigated as general-purpose optimizers~\citep{yang2023large,meyerson2024language,liu2024large} and extended to this domain through prompt-based methods that propose modifications using external databases~\citep{wang2024efficient,liu2024conversational}. 
To enhance generation quality, other frameworks leverage representation learning~\citep{rankovic2023bochemian,bedrosian2024small,fang2023domain,kristiadi2024sober} or instruction tuning on large-scale datasets~\citep{ye2025drugassist,nguyen2024lico}. 
Despite these advances, most existing systems still directly generate final molecular representations from high-level objectives or reasoning, while the interface between high-level molecular reasoning and low-level structural manipulation remains underdeveloped. \abbr{} bridges this gap by introducing code as an execution interface between chemical reasoning and molecular graphs, enabling structured editing intentions to be translated into consistent, verifiable, and controllable molecular modifications.
\section{Methods}

\subsection{Problem formulation}

Molecular optimization is a central task in drug discovery, where the objective is to generate a modified molecule $M_o$ from an initial compound $M_i$ to improve one or more target properties $T$ (e.g., permeability, target binding affinity), while preserving essential structural features such as the core scaffold or pharmacophores, and enabling reuse of steps in a common synthetic route.

In many prior approaches, particularly those based on RNNs or early LLMs, this task is formulated as:
\begin{equation}
    M_o = \mathcal{F}(M_i, T)
\end{equation}
where $\mathcal{F}$ is a black-box model that directly maps the input molecule and target to a SMILES string.

While this formulation has shown some empirical success, it suffers from two key limitations. First, the editing process is entirely implicit: the model does not explain what was changed or why, making the transformation uninterpretable. Second, the output often diverges from well-established medicinal chemistry principles. These models tend to make broad, unconstrained modifications, rather than the minimal, targeted edits that chemists use, such as modifying a functional group to adjust polarity for better permeability, or introducing an H-bond donor to improve affinity and selectivity. Without such structure–property reasoning, the results are hard to interpret or trust, limiting their practical utility in real-world design workflows.

In contrast, recent reasoning-centric LLMs offer a fundamentally different capability: rather than directly generating $M_o$, they can first articulate a structured editing intention based on $M_i$ and $T$. These intentions often consist of: \textbf{a set of editing actions} $A = \{a_1, a_2, \dots\}$ (e.g., replace the para-methyl with a hydroxyl group), and \textbf{a corresponding rationale} (e.g., to increase polarity and improve solubility).

\begin{figure*}[t]
\begin{center}
\includegraphics[width=1.\textwidth]{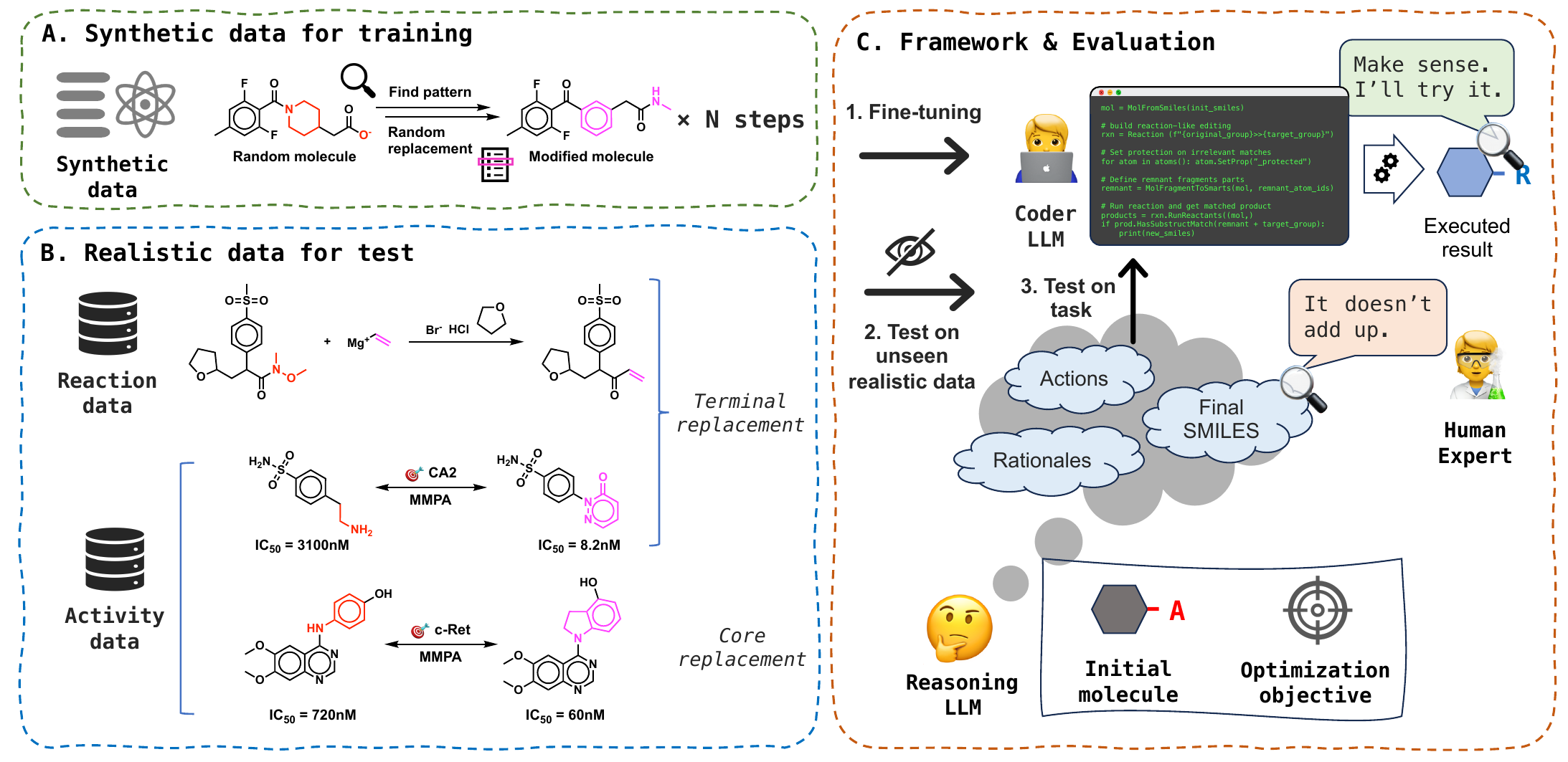}
\end{center}
\caption{Overview of the \abbr{} framework. 
(A) Synthetic data construction for model training.
(B) Realistic data for testing on unseen molecular modifications derived from both reaction and activity data, reflecting chemical transformations observed in practice and in human reasoning. Modifications are categorized into terminal or core replacements. 
(C) End-to-end workflow and evaluation of the framework. The coder LLM is evaluated both independently and within the integrated pipeline. A reasoning LLM first generates rationales and editing actions, which are then executed by the coder model to produce optimized molecules. Outputs are subsequently reviewed by human experts to assess the alignment between generated structures and intended editing actions.}
\label{fig:data_framework}
\end{figure*}

This shift from black-box generation to interpretable, rationale-driven editing actions more closely reflects how medicinal chemists reason about molecular design based on structural and physicochemical considerations. The corresponding formulation becomes:
\begin{equation}
    (A, M_o) \leftarrow \mathcal{F}(M_i, T)
\end{equation}
However, in current systems, this promising capability remains underutilized. While reasoning LLMs can propose chemically meaningful edits, they often fail to execute them faithfully. This is largely due to the difficulty of performing precise and constrained modifications on SMILES representations, which are sensitive to minor syntax errors and lack structural locality. As a result, the generated molecule $M_o$ may deviate from the proposed intention or violate chemical constraints.

To bridge this gap between reasoning and execution, we reformulate molecular editing as a code generation task:
\begin{equation}
\mathcal{C} = \mathcal{G}(M_i, A),\quad M_o = \mathcal{C}(M_i)
\end{equation}
where $\mathcal{G}$ denotes the code generation model (e.g., a coder LLM), and $\mathcal{C}$ is an executable script (e.g., using RDKit) that implements the specified editing actions $A$. 

\paragraph{Why code as an interface?}
LLMs have demonstrated strong performance in translating natural language into structured domain-specific languages such as programming code~\citep{fang_llm4code-eval_2024}. Models like CodeT5+~\citep{wang2023codet5+} and StarCoder~\citep{li2023starcoder} can reliably generate and edit code given natural language instructions. This success is attributed to the syntactic and semantic regularity of code, and structure-aware strategies such as abstract syntax trees~\citep{wang_codet5_2021}, fill-in-the-middle training~\citep{li2023starcoder}, and execution-based feedback~\citep{li_alphacode_2022}.

This motivates us to treat molecular editing as a code generation task. Rather than generating fragile SMILES strings, the LLM produces executable scripts that specify structural edits. Notably, cheminformatics libraries like RDKit~\citep{landrum_rdkit_2013} provide robust APIs for manipulating molecular graphs, allowing SMILES strings to be parsed into graph-based data structures and modified programmatically. This enables verifiable and reproducible molecular transformations grounded in chemical structure, not string syntax.

\subsection{Framework overview}

We propose \textbf{\abbr} (\textbf{M}olecular \textbf{E}diting via \textbf{Co}de generation), a framework that decouples high-level chemical reasoning from low-level molecular editing, as illustrated in Figure~\ref{fig:data_framework}C. Given an initial molecule $M_i$ and a target property goal $T$, a reasoning LLM first generates an editing intention $(A, \text{rationale})$, where $A$ denotes a set of editing actions and the rationale explains their purpose. The focus shifts to executing $A$ through code generation, producing the optimized molecule $M_o$.

\abbr{} operates in a cascading manner:

\textbf{Intention generation (reasoning LLM):} Generates editing actions $A$ and associated rationales based on $(M_i, T)$.

\textbf{Action execution (code LLM):} Translates $A$ into code $\mathcal{C}$, which is applied to $M_i$ to yield $M_o$.
\begin{equation}
\begin{aligned}
\underbrace{(M_i, T) 
\longrightarrow (A,~\text{rationale})}_{\text{reasoning LLM}}
&\longrightarrow 
\underbrace{\mathcal{C} = \text{CodeGen}(A)}_{\text{code LLM}} \\
&\longrightarrow 
M_o = \mathcal{C}(M_i)
\end{aligned}
\end{equation}
This decoupled formulation offers several advantages:
(1) It separates reasoning (\textit{why} to edit) from execution (\textit{how} to edit), mirroring expert design workflows.
(2) It provides interpretable, verifiable, and debuggable records of molecular transformation.
(3) It ensures structural validity and fine-grained control by grounding edits in explicitly defined procedures.

By translating natural language intentions into code-level structural edits, \abbr{} enables LLM-driven molecular design that is more reliable, transparent, and amenable to practical drug-discovery workflows.

\subsection{Data construction}

To train the code LLM to translate editing actions into executable molecular transformations, we construct a large-scale dataset of molecule–edit–code triples in the form:
\begin{equation}
(M_i, A) \longrightarrow \mathcal{C}
\end{equation}
where $M_i$ is the input molecule, $A$ is a set of editing actions expressed in natural language, and $\mathcal{C}$ is the corresponding Python code that applies the edit using cheminformatics toolkits such as RDKit.

We adopt a hybrid data construction strategy combining synthetic and realistic samples:

\textbf{Synthetic edits.}  \label{para:syn-edits}
We constructed synthetic editing examples by applying programmatically defined moiety replacements to molecules randomly sampled from the ZINC database~\citep{sterling_zinc_2015}, as shown in Figure~\ref{fig:data_framework}A. Each initial molecule $M_i$ was iteratively edited by substituting one of its substructures with a moiety from a predefined pattern pool $\mathcal{P} = \bigcup_i \mathcal{P}_i$, where each subset $\mathcal{P}_i$ contains fragments with $i$ attachment points.

This procedure yields chemically valid and structurally controlled synthetic edits, providing precise edit-code supervision that enables the model to learn robust mappings from actions to code without requiring exposure to real-world editing data. The full method and list of predefined patterns are provided in Appendix~\ref{app:synthetic-edits}.

\textbf{Realistic edits.} \label{para:data-construction}
To evaluate generalization beyond synthetic edits, we construct a set of realistic editing samples derived from two sources (Figure~\ref{fig:data_framework}B, detailed in Appendix~\ref{app:real-edits}): (1) Reaction-derived edits, where we retained samples with a single modification site while preserving the molecular core. (2) Target-based edits, where we extracted target-specific compounds from the ChEMBL35 database and processed with matched molecular pairs analysis. The data were categorized into two types: \textbf{terminal replacements} (single-point attachment, same as those in \textbf{Reaction-derived edits}) and \textbf{core replacements} (multi-point attachment).

Together, these two sources reflect the two most common strategies in real-world molecular design. The first captures reaction-derived transformations, grounded in feasible chemical synthesis steps. The second comprises target-specific structural modifications observed across bioactive compounds, which represent plausible edits made during lead optimization. While the directionality of property change is not explicitly defined in our setup, these samples provide chemically meaningful, human-curated edits. Although our model is trained solely on synthetic data, evaluation on these realistic edits allows us to assess its generalization to authentic molecular transformation scenarios. 

\subsection{Model Training}

To train the coder LLM for accurate molecular editing, we used a curated dataset of 50,000 synthetic editing samples (Section~\ref{para:syn-edits}). Each sample includes a molecule $M_i$, an editing action $A$ described using a combination of natural language and cheminformatics notation (e.g., SMARTS), and a Python code snippet $\mathcal{C}$ that transforms $M_i$ accordingly. For training the direct SMILES generation baseline, the corresponding output molecules $M_o = \mathcal{C}(M_i)$ were used as targets. All models were fine-tuned using the official Qwen2.5-Coder finetuning framework. Additional details regarding prompt design, formatting choices, and training settings are provided in Appendix~\ref{app:model-training}.

\subsection{Benchmark and evaluation metrics}

To evaluate both the core editing capability and downstream utility of our framework, we introduced two complementary benchmarks aligned with our problem formulation:

\textbf{1. Molecular editing benchmark for problem diagnosis.}
We used the realistic editing samples constructed in Section~\ref{para:data-construction} to identify the limitations of existing approaches and to quantitatively evaluate the molecular editing capabilities of \abbr{}. For each sample, we assess whether the model can faithfully execute the specified editing action $A$ on the input molecule $M_i$, either directly or via code generation, to produce the expected output molecule $M_o$.

\begin{table*}[t]
\centering

\resizebox{0.85\textwidth}{!}{%
\begin{tabular}{llccc}
\toprule
\multirow{2}{*}{\textbf{Action}} & \multirow{2}{*}{\textbf{Model}} 
& \multicolumn{2}{c}{\textbf{Terminal replacement}} 
& \textbf{Core replacement} \\
\cmidrule(lr){3-4} \cmidrule(lr){5-5}
& & \textbf{Reaction-derived} & \textbf{Target-specific} & \textbf{Target-specific}  \\
\midrule
\multirow{2}{*}{Direct} 
& \cellcolor{blue!10} Qwen2.5-7B-Instruct & \cellcolor{blue!10} 0.5 & \cellcolor{blue!10} 1.3 & \cellcolor{blue!10} 0.0 \\
& \cellcolor{blue!20} Qwen2.5-7B-Instruct-FT & \cellcolor{blue!20} 38.7 & \cellcolor{blue!20} 7.4 & \cellcolor{blue!20} 2.7 \\
\midrule
\multirow{4}{*}{CodeGen} 
& \cellcolor{blue!10} Qwen2.5-7B-Instruct & \cellcolor{blue!10} 35.4 & \cellcolor{blue!10} 3.7 & \cellcolor{blue!10} 4.1 \\
& \cellcolor{green!10} Qwen2.5-Coder-7B-Instruct & \cellcolor{green!10} 50.3 & \cellcolor{green!10} 9.9 & \cellcolor{green!10} 12.2 \\
& \cellcolor{green!20} Qwen2.5-Coder-7B-Instruct-FT (ours) & \cellcolor{green!20} \textbf{99.9} & \cellcolor{green!20} \textbf{98.3} & \cellcolor{green!20} \textbf{98.3} \\
\bottomrule
\end{tabular}%
}
\caption{
Execution accuracy (\%) across realistic edit benchmarks.
Rows are shaded to distinguish model variants:
\colorbox{blue!10}{Qwen2.5},
\colorbox{green!10}{Qwen2.5-Coder},
and \colorbox{blue!20}{finetuned} \colorbox{green!20}{versions} (darker, denoted by -FT).
}
\label{tab:realistic-edit-accuracy}
\end{table*}

\begin{table*}[t]
\centering
\resizebox{0.88\textwidth}{!}{%
\begin{tabular}{lcccccccccccc}
\toprule
\multicolumn{13}{c}{\textbf{Property optimization tasks}} \\
\midrule
\multirow{2}{*}{\textbf{Model}} &
\multicolumn{4}{c}{\textbf{Penalized logP}} &
\multicolumn{4}{c}{\textbf{Solubility}} &
\multicolumn{4}{c}{\textbf{QED}} \\
\cmidrule(lr){2-5} \cmidrule(lr){6-9} \cmidrule(lr){10-13}
& VR\% & SR\% & Sim & CR\%
& VR\% & SR\% & Sim & CR\%
& VR\% & SR\% & Sim & CR\% \\
\midrule
\multicolumn{13}{l}{\textit{W/o thinking}} \\
\midrule
GPT-4o           & 22 & 8  & 0.11 &  --   & 41 & 35 & 0.21 &   --  & 31 & 17 & 0.11 & --    \\
Gemini-2.0-flash & 66 & 44 & 0.27 &   --  & 32 & 28 & 0.13 &  --   & 64 & 54 & 0.26 & --    \\
DeepSeek-V3      & 53 & 21 & 0.20 &   --  & 52 & 44 & 0.21 &  --   & 48 & 33 & 0.18 &  --   \\
\midrule
\multicolumn{13}{l}{\textit{W/ thinking}} \\
\midrule
GPT-5             & 58 & 32 & 0.21 &  --   & 67 & 63 & 0.25 &  --   & 68 & 66 & 0.21 &   --  \\
Claude-Sonnet-4.6 & 71 & 67 & 0.28 & -- & 66 & 66 & 0.28 & -- & 74 & 70 & 0.32 & -- \\
Gemini-2.5-Pro    & 78 & 69 & 0.37 &  45   & 79 & 79 & 0.40 &   42  & 83 & \textbf{83} & 0.33 &   55  \\
DeepSeek-R1       & 62 & 48 & 0.31 & 30  & 70 & 64 & 0.39 & 28  & 68 & 61 & 0.31 & 29  \\
\midrule
\textbf{\abbr{}}     & \textbf{93} & \textbf{72} & \textbf{0.54} & \textbf{96}  & \textbf{96} & \textbf{96} & \textbf{0.63} & \textbf{95}  & \textbf{87} & 75 & \textbf{0.49} & \textbf{95} \\
\midrule
\multicolumn{13}{c}{\textbf{Activity optimization tasks}} \\
\midrule
\multirow{2}{*}{\textbf{Model}} &
\multicolumn{4}{c}{\textbf{DRD2}} &
\multicolumn{4}{c}{\textbf{JNK3}} &
\multicolumn{4}{c}{\textbf{GSK3$\beta$}} \\
\cmidrule(lr){2-5} \cmidrule(lr){6-9} \cmidrule(lr){10-13}
& VR\% & SR\% & Sim & CR\%
& VR\% & SR\% & Sim & CR\%
& VR\% & SR\% & Sim & CR\% \\
\midrule
\multicolumn{13}{l}{\textit{W/o thinking}} \\
\midrule
GPT-4o           & 27 & 12 & 0.12 &   --  & 20 & 4  & 0.08 &   --  & 19 & 10 & 0.08 & --    \\
Gemini-2.0-flash & 60 & 32 & 0.29 &   --  & 56 & 17 & 0.25 &  --   & 71 & 39 & 0.36 &  --   \\
DeepSeek-V3      & 46 & 22 & 0.20 &   --  & 40 & 9  & 0.14 &  --   & 43 & 18 & 0.15 & --    \\
\midrule
\multicolumn{13}{l}{\textit{W/ thinking}} \\
\midrule
GPT-5             & 64 & 50 & 0.20 &  --   & 57 & 15 & 0.17 &   --  & 46 & 28 & 0.14 &   --  \\
Claude-Sonnet-4.6 & 67 & 53 & 0.28 & -- & 39 & 9 & 0.13 & -- & 56 & 40 & 0.20 & -- \\
Gemini-2.5-Pro    & 85 & 64 & 0.36 &  43   & 74 & 24 & 0.32 &  42   & 74 & 51 & 0.38 &  29   \\
DeepSeek-R1       & 73 & 54 & 0.36 & 33  & 43 & 8  & 0.19 & 12  & 45 & 28 & 0.22 & 20  \\
\midrule
\textbf{\abbr{}}     & \textbf{89} & \textbf{66} & \textbf{0.55} & \textbf{93}  & \textbf{92} & \textbf{39} & \textbf{0.55} & \textbf{98}  & \textbf{91} & \textbf{60} & \textbf{0.56} & \textbf{96} \\
\bottomrule
\end{tabular}%
}%
\caption{
Performance on molecular optimization tasks. Each task reports validity rate (VR\%), success rate (SR\%), mean similarity to the source molecule (Sim), and consistency rate (CR\%). 
\textbf{Bold} numbers highlight the best value in each column.
Note: GPT-5 refused to answer a small number of samples.
}
\label{tab:molopt-tasks}
\end{table*}

\textbf{2. Molecular modification benchmark for application endpoint.}
To assess the practical utility of our framework, we evaluate end-to-end performance on ChemCoTBench~\citep{li_ChemCoTBench_2025}. This benchmark comprises six molecular modification tasks: three involving physicochemical properties and three involving target-specific bioactivities. We compare direct molecule generation from a reasoning LLM with the MECo framework in terms of (1) chemical validity, (2) optimization success rate, (3) mean similarity to source molecules, and (4) structure–action consistency.

\begin{figure*}[t]
\begin{center}
\includegraphics[width=0.95\textwidth]{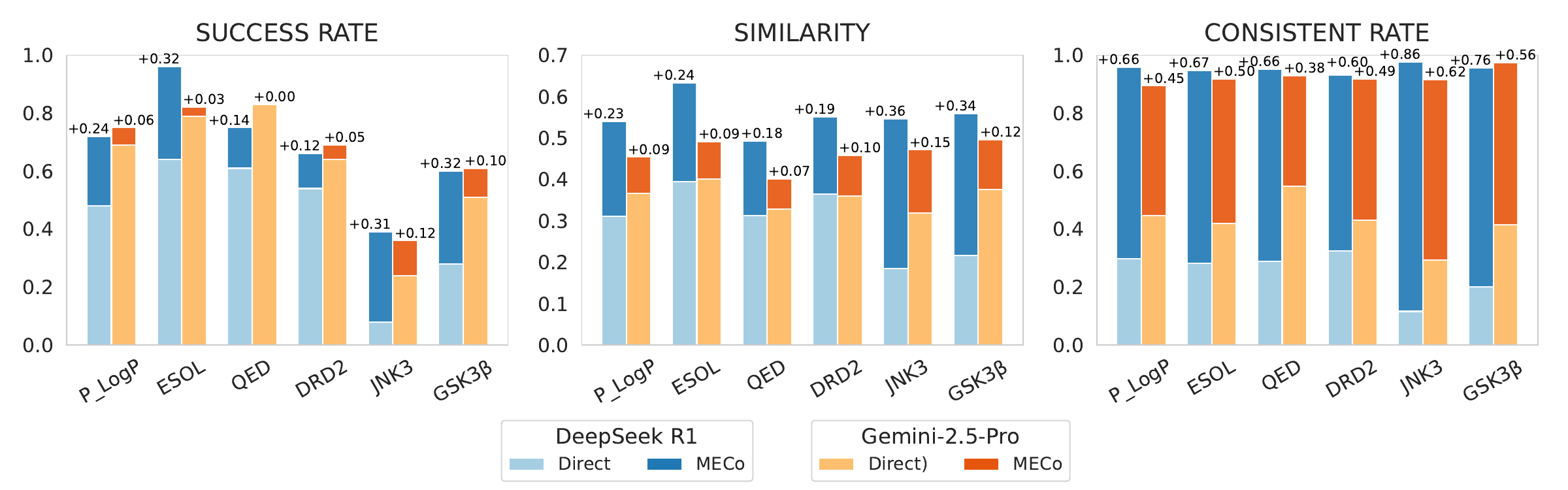}
\end{center}
\caption{Incremental bar plots comparing direct generation (light bars) and \abbr{} (dark stacked bars) across tasks and metrics under two different reasoning LLMs. Numbers above the bars indicate the absolute improvements of \abbr{} over direct.}
\label{fig:\abbr{}_improv}
\end{figure*}

\begin{figure*}[t]
\begin{center}
\includegraphics[width=1.\textwidth]{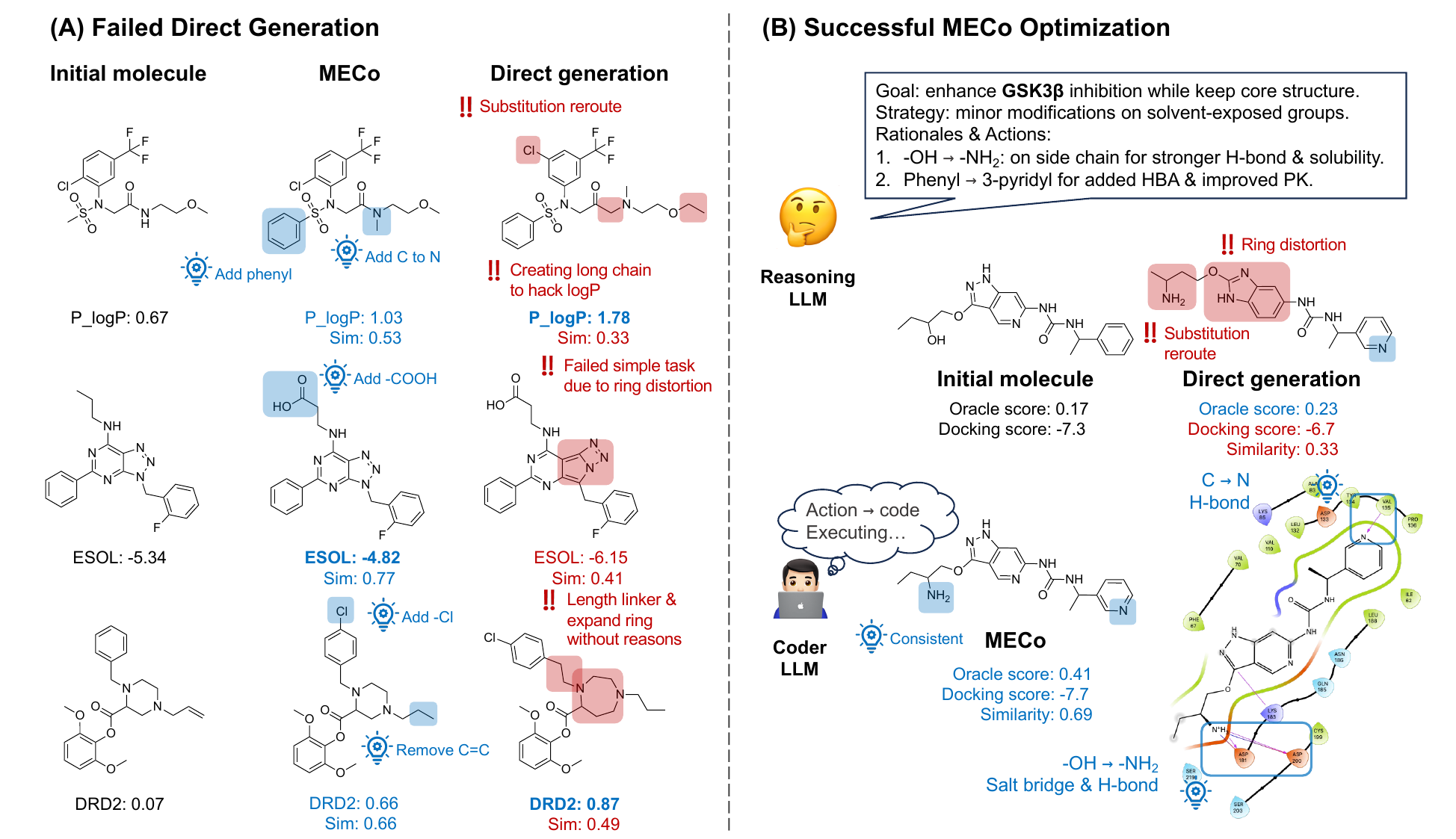}
\end{center}
\caption{Representative case study of molecular modification. (A) Direct generation produces uncontrolled changes and low similarity, often leading to unrealistic score inflation or failed optimization; \abbr{} achieves improved scores and higher similarity through interpretable, chemistry-preserving edits. (B) \abbr{} demonstrates coherent consistency across the reasoning–execution–analysis pipeline.}
\label{fig:case}
\end{figure*}

\section{Experiments}

\subsection{Experimental setup}

In the molecular modification task under the \abbr{} framework, we applied a unified wrapper to the original prompt to constrain the output format and facilitate reliable extraction of editing actions. The full wrapper template is provided in Appendix~\ref{app:prom_wrap}. To obtain ground-truth labels, we manually applied the actions generated by the reasoning LLMs to the initial molecules, following the evaluation criteria described in Appendix~\ref{app:manual_criteria}.

We selected DeepSeek-R1~\citep{guo_deepseek-r1_2025} as the reasoning LLM used in \abbr{} by default, due to its strong performance in structured scientific reasoning and its fully open-source availability, which facilitates reproducibility and downstream integration. Gemini-2.5-Pro~\citep{comanici_gemini_2025} was also evaluated for comparison, given its top performance on ChemCoTBench.
We employed Qwen2.5-7B~\citep{qwen_qwen25_2025} to showcase an open-source implementation of the edit execution model, fine-tuning its Coder variant~\citep{hui2024qwen2} for code-based editing, and its general-purpose variant for direct SMILES generation.

\subsection{Edit execution on realistic benchmark} \label{subsec:edit_real}

To ensure a fair evaluation of editing generalization, we removed all samples from each realistic dataset whose original or modified fragments had a Tanimoto similarity $\geq 0.6$ (computed using ECFP4~\citep{rogers_extended-connectivity_2010}) with any moiety used in the synthetic training set. From the remaining pool, we randomly sampled 1,000 examples per category: reaction-derived, target-specific terminal replacement, and target-specific core replacement, as the test sets. This setup ensures that the model is evaluated on structurally diverse and unseen edits, providing a robust assessment of its generalization ability. Results are summarized in Table~\ref{tab:realistic-edit-accuracy}. Appendix Figure~\ref{fig:frag-tsne} further visualizes the fragment distributions via t-SNE, showing the broad and largely non-overlapping chemical space of realistic edits.

This comparison demonstrates the advantage of code-based molecular editing over direct SMILES generation, independent of both string and structural similarities (see Appendix Figure~\ref{fig:acc_sim_bins}). The direct generation model Qwen2.5-7B-Instruct exhibits extremely low execution accuracy across all categories, suggesting a limited ability to faithfully apply structural modifications. Notably, even without specialized training, prompting the same general LLM to generate code leads to substantially higher accuracy than its direct generation counterpart, underscoring the inherent advantages of structured code over direct SMILES generation. Interestingly, we find that the discrepancy between reaction-derived and target-specific terminal replacements can be largely attributed to differences in SMILES syntax patterns, suggesting that LLMs struggle to generalize over SMILES syntax in both direct generation and code generation. A detailed discussion is provided in Appendix~\ref{app:ob_ter_re}.

With supervised fine-tuning, the code-specialized model Qwen2.5-Coder-7B-Instruct-FT achieves over 98\% execution accuracy across all benchmarks, while direct generation shows only marginal improvement. These results affirm that \abbr{}'s code generation formulation dramatically improves execution fidelity, even on realistic and diverse editing tasks.

\subsection{Improvement on molecular modification}

Table~\ref{tab:molopt-tasks} summarizes performance across six molecular optimization tasks, covering both physicochemical properties and target activities. We compare several non-reasoning / reasoning LLMs with our proposed \abbr{} framework, which augments reasoning-centric molecular modification with code generation for edit execution. We omit mean property improvement from the main results, since models can achieve large gains by sacrificing structural preservation and producing molecules that differ substantially from the initial ones. Such cases inflate the average improvement without reflecting meaningful optimization. We therefore adopt success rate, and provide the mean improvement results in Appendix~\ref{app:imp_sim}. 


\abbr{} achieves markedly higher consistency rates (CR\%), indicating that the resulting structures more faithfully realize the editing intentions proposed by the reasoning LLM. This closer intention--structure alignment improves the transparency and interpretability of the editing process by making the intended modifications explicit and their structural realization directly verifiable. Together with the gains in validity, success rate, and structural similarity, these results suggest that more faithful execution helps translate editing strategies proposed by the reasoning LLMs into valid, successful, and structurally conservative molecular modifications.

In contrast, direct SMILES generation often produces unintended or overly disruptive modifications, with poor alignment to the specified editing instructions (see cases in Section~\ref{sec:case}). Even strong reasoning LLMs fall short when lacking explicit action execution, underscoring the need to bridge high-level reasoning and low-level structural manipulation through code-based editing.

Figure~\ref{fig:\abbr{}_improv} further shows that \abbr{}’s improvements generalize across reasoning LLMs (DeepSeek-R1 and Gemini-2.5-Pro), demonstrating its broad applicability and utility in controllable, interpretable molecule design.


\subsection{Case study} \label{sec:case}

To illustrate the difference between direct generation and our \abbr{} framework, we analyze molecular modification tasks and present representative cases in Figure~\ref{fig:case}A. Direct generation often introduces uncontrolled structural changes with low similarity to the initial molecule, leading to unrealistic score inflation or failed optimization. In contrast, \abbr{} achieves improved scores while maintaining high similarity through interpretable, chemistry-preserving edits.

The detailed example on the GSK3$\beta$ activity optimization task in Figure~\ref{fig:case}B further demonstrates how \abbr{} maintains consistency across the reasoning–execution–analysis pipeline. The reasoning LLM proposes rational modifications, such as replacing a hydroxy with an amino to strengthen hydrogen bonding and improve solubility, and substituting a phenyl ring with a pyridin-3-yl ring to introduce an additional hydrogen bond acceptor and enhance pharmacokinetic potential. The coder LLM then faithfully translates these actions into executable edits, ensuring that the intended modifications are precisely applied. This yields improved oracle and docking scores, while complementary binding analysis further supports the rationales. The coherent consistency underpins the interpretability and reliability of \abbr{} compared to direct generation.
\section{Conclusion}

We introduce \abbr{}, a framework that reformulates molecular modification as a code generation task, bridging high-level reasoning with low-level structural execution. By translating interpretable editing intentions into executable scripts, \abbr{} achieves near-perfect edit fidelity on realistic benchmarks and consistently outperforms direct-SMILES-generation baselines across both physicochemical property and bioactivity optimization tasks. This formulation enables consistent, controllable and interpretable molecular modification workflows, paving the way for reliable human-AI collaboration and future high-fidelity feedback loops in iterative molecular design workflows, such as agent-style extensions and multi-round optimization. Beyond this application, our results suggest a general paradigm for bridging reasoning and execution in scientific domains, highlighting the potential of LLMs as reliable assistants for structured, verifiable discovery workflows.

\section*{Limitations}

Our work decouples molecular reasoning from structural execution, but its end-to-end performance remains bounded by the quality of the upstream reasoning LLM: if the reasoning model proposes chemically plausible but unreasonable editing actions, the coder LLM will execute them faithfully rather than correct them. This propagation of upstream errors is an inherent consequence of the cascaded design.

While multi-step replacements are sufficient to express virtually any structural modification at the molecular level, certain medicinal chemistry design strategies—such as ring opening/closure and bond rerouting—can be somewhat unintuitive when represented purely as replacements. Extending the framework to more naturally accommodate these transformations could better align model representations with expert medicinal chemistry priors, potentially improving the utilization of domain knowledge and the quality of generated modifications, which is important for practical deployment and system optimization.

The consistency rate (CR\%) is a metric specifically designed to evaluate intention--execution fidelity and is not intended as an independent measure of molecular or pharmacological optimization quality. Details on its manual evaluation, subjectivity, and scalability are provided in Appendix~\ref{app:manual_criteria}.

\section*{Acknowledgments}

This work is supported by Innovative Drug Research
and Development-National Science and Technology Major Project (No.2025ZD1802501) and Beijing Frontier Research Center for Biological Structure Fundings.

\bibliography{example_paper}



\newpage
\appendix

\section{Data construction details}
\subsection{Synthetic edit generation}
\label{app:synthetic-edits}

We provide here the full algorithm for generating synthetic editing examples, as described in Section~\ref{para:syn-edits}. The procedure applies pre-defined moiety replacements to randomly sampled molecules from the ZINC database, using structured pattern pools grouped by attachment point count.

\begin{algorithm}
\caption{Iterative Moiety Replacement}
\begin{algorithmic}[1]
\REQUIRE Molecule $M_i$, pattern pool $\mathcal{P} = \bigcup_i \mathcal{P}_i$, iterations $N$
\STATE $M \gets M_i$ \COMMENT{Initialize molecule}
\FOR{$n = 1$ to $N$}
    \STATE $\mathcal{P}_\text{avail} \gets \mathcal{P}$ \COMMENT{Reset available patterns}
    \STATE $\mathit{replaced} \gets \text{False}$
    \WHILE{$\mathcal{P}_\text{avail} \ne \emptyset$ and $\mathit{replaced} = \text{False}$}
        \STATE Sample $p \sim \mathcal{P}_\text{avail}$; $\mathcal{P}_\text{avail} \gets \mathcal{P}_\text{avail} \setminus \{p\}$ \COMMENT{Draw and remove pattern}
        \STATE $\mathcal{S} \gets \textsc{GetSubstructMatches}(M, p)$ \COMMENT{Identify all matching substructures}
        \IF{$\mathcal{S} \ne \emptyset$}
            \STATE $s \sim \mathcal{S}$ \COMMENT{Randomly select a match site}
            \STATE $i \gets \text{attachment\_count}(p)$ \COMMENT{Determine connection type}
            \STATE $r \sim \mathcal{P}_i \setminus \{p\}$ \COMMENT{Sample replacement with same connectivity}
            \FOR{$a$ in $M$}
                \IF{$a \notin s$}
                    \STATE Protect($a$) \COMMENT{Protect atoms not in the selected substructure match $s$}
                \ENDIF
            \ENDFOR
            \STATE $M \gets \text{replace}(M, p, r)$ \COMMENT{Apply replacement using \textsc{Reaction}}
            \STATE Record $(p, s, r)$ \COMMENT{$s$ uniquely specifies the matched substructure instance}
            \STATE $\mathit{replaced} \gets \text{True}$
        \ENDIF
    \ENDWHILE
    \IF{$\mathit{replaced} = \text{False}$}
        \STATE \textbf{break} \COMMENT{Stop further iterations if replacement failed}
    \ENDIF
\ENDFOR
\RETURN $M_o,\, \{(p_j, s_j, r_j)\}_{j=1}^n$ \COMMENT{Final molecule and list of edits}
\end{algorithmic}
\end{algorithm}

At each iteration, a pattern $p$ was randomly drawn from the pool, and a matching site in $M_i$ was identified. If a match was found, it was replaced with another randomly selected moiety $r \in \mathcal{P}_i \setminus \{p\}$ of the same connectivity. If no match was found, the process continued with a different pattern until the pool was exhausted. Replacement was performed iteratively, with all successful edits recorded. For each edit, the corresponding executable code snippet $\mathcal{C}$ was automatically generated, and the edited molecule $M_o$ was obtained by applying $\mathcal{C}$ to the input molecule $M_i$, i.e., $M_o = \mathcal{C}(M_i)$.

The lists of predefined substituents (Table~\ref{stab:rgroup}) and molecular linkers (Table~\ref{stab:linker}) are provided for constructing synthetic editing samples.

\begin{table*}[t]
\centering
\begin{tabular}{lll}
\toprule
Category & Name & SMILES \\
\midrule
\multirow{4}{*}{Halogens}
  & Fluoro  & [*:1]F \\
  & Chloro  & [*:1]Cl \\
  & Bromo   & [*:1]Br \\
  & Iodo    & [*:1]I \\
\midrule
\multirow{4}{*}{Alkyl}
  & Methyl       & [*:1]C \\
  & Ethyl        & [*:1]CC \\
  & Isopropyl    & [*:1]C(C)C \\
  & tert-Butyl   & [*:1]C(C)(C)C \\
\midrule
\multirow{3}{*}{Aryl}
  & Phenyl           & [*:1]c1ccccc1 \\
  & p-Tolyl          & [*:1]c1ccc(cc1)C \\
  & p-Chlorophenyl   & [*:1]c1ccc(cc1)Cl \\
\midrule
\multirow{6}{*}{Oxygen-containing}
  & Hydroxyl   & [*:1]O \\
  & Methoxy    & [*:1]OC \\
  & Ethoxy     & [*:1]OCC \\
  & Carboxyl   & [*:1]C(=O)O \\
  & Aldehyde   & [*:1]C=O \\
  & Ketone     & [*:1]C(=O)C \\
\midrule
\multirow{5}{*}{Nitrogen-containing}
  & Amino          & [*:1]N \\
  & Methylamino    & [*:1]NC \\
  & Dimethylamino  & [*:1]N(C)C \\
  & Cyano          & [*:1]C\#N \\
  & Nitro          & [*:1][N+](=O)[O-] \\
\midrule
\multirow{3}{*}{Sulfur-containing}
  & Thiol      & [*:1]S \\
  & Methylthio & [*:1]SC \\
  & Sulfonyl   & [*:1]S(=O)(=O)C \\
\bottomrule
\end{tabular}
\caption{Common substituents for synthetic data.}
\label{stab:rgroup}
\end{table*}

\begin{table*}[t]
\centering
\begin{tabular}{lll}
\toprule
Category & Name & SMILES \\
\midrule
\multirow{3}{*}{Aromatic linkers}
  & Meta-phenylene  & [*:1]c1cc([*:2])ccc1 \\
  & Para-phenylene  & [*:1]c1ccc([*:2])cc1 \\
  & Ortho-phenylene & [*:1]c1c([*:2])cccc1 \\
\midrule
\multirow{7}{*}{Carbonyl-based linkers}
  & Amide          & [*:1][C;!R](=O)[N;!R][*:2] \\
  & Reverse amide  & [*:1][N;!R][C;!R](=O)[*:2] \\
  & Ester          & [*:1][C;!R](=O)[O;!R][*:2] \\
  & Ketone bridge  & [*:1][C;!R](=O)[*:2] \\
  & Urea           & [*:1][N;!R][C;!R](=O)[N;!R][*:2] \\
  & Carbamate      & [*:1][O;!R][C;!R](=O)[N;!R][*:2] \\
  & Sulfonamide    & [*:1]S(=O)(=O)[N;!R][*:2] \\
\midrule
\multirow{5}{*}{Alkyl / heteroatom linkers}
  & Methylene        & [*:1][C;!R][*:2] \\
  & Ethylene         & [*:1][C;!R][C;!R][*:2] \\
  & Ether            & [*:1][O;!R][*:2] \\
  & Thioether        & [*:1][S;!R][*:2] \\
  & Secondary amine  & [*:1][N;!R][*:2] \\
\midrule
\multirow{4}{*}{Extended / heterocyclic linkers}
  & 1,2,3-Triazole   & [*:1]c1nnn([*:2])c1 \\
  & Imidazole-type   & [*:1]c1[nH]cc([*:2])n1 \\
  & Piperazine       & [*:1]N1CCN([*:2])CC1 \\
  & Piperidine       & [*:1]N1CCC([*:2])CC1 \\
\midrule
Polar chain linker
  & PEG unit (ethylene glycol) & [*:1][O;!R][C;!R][C;!R][O;!R][*:2] \\
\bottomrule
\end{tabular}
\caption{Common molecular linkers for synthetic data.}
\label{stab:linker}
\end{table*}

\subsection{Realistic edit generation}
\label{app:real-edits}

We construct a set of realistic editing samples derived from two sources:

(1) Reaction-derived edits: We extracted matched reactant–product pairs from the USPTO-MIT reaction dataset~\citep{jin_uspto_2017}. Atom mapping was used to identify structural changes, and we retained samples with a single modification site while preserving the molecular core.

(2) Target-based edits: We extracted compounds from the ChEMBL35 database~\citep{mendez_chembl_2018} and grouped them by target ID after applying basic structure-based filters (details provided in Appendix~\ref{app:data-filtering}). Matched molecular pairs (MMPs) were identified using the algorithm by \citet{hussain_mmp_2010}, and further filtered following the criteria described in Appendix~\ref{app:mmp-filtering}. Attachment points were explicitly identified and recorded to enable precise modification tracking. Based on the number of attachment points in each transformation, the data were categorized into two types: \textbf{terminal replacements} (single-point attachment, same as those in \textbf{Reaction-derived edits}) and \textbf{core replacements} (multi-point attachment).

\subsection{Compound preprocessing and filtering}
\label{app:data-filtering}

We extracted standardized SMILES strings and associated bioactivity data from the ChEMBL35 database. To ensure molecular quality and downstream compatibility, we applied the following filtering criteria:
\begin{itemize}
    \item Salt removal: only the largest fragment was retained;
    \item Molecular weight between 100 and 800 Da;
    \item Allowed atom types: H, C, N, O, F, Cl, Br, I, S, P, B, Se;
    \item No linear unbranched chains longer than six heavy atoms.
\end{itemize}

These filters resulted in a curated set of drug-like molecules, which served as the input for MMP generation.

\subsection{Target grouping and MMP Analysis} \label{app:mmp-filtering}

After filtering, compounds were grouped by target ID, and each group was exported as a separate \texttt{.smi} file. For each group, matched molecular pairs (MMPs) were generated using the RDKit Contrib MMPA pipeline (\texttt{rfrag.py} and \texttt{indexing.py}). This procedure applies systematic bond fragmentation and indexing to identify molecular pairs that differ by a single localized transformation. Each transformation was encoded as a SMIRKS pattern, representing the minimal structural change. All MMP results were merged and formed a high-quality pool of target-specific matched molecular pairs.

\subsection{Transformation type classification}

To better characterize the nature of molecular edits, the resulting MMPs were further categorized into two types based on structural properties:

\paragraph{1. Core replacement.}  
MMPs were classified as core replacements if they satisfied the following criteria:
\begin{itemize}
    \item \textbf{Attachment point constraint:} both fragments have the same number of attachment points, with count $\ge2$.
    \item \textbf{Scaffold difference:} the Murcko scaffolds of the prior and latter fragments in replacement are different;
    \item \textbf{Ring requirement:} both fragments contain at least one ring;
    \item \textbf{Non-ring atom constraint:} each fragment contains no more than 5 non-ring heavy atoms, including the attachment points, to avoid heavily decorated rings;
    \item \textbf{Component ratio:} the core fragment in the initial molecule accounts for less than 50\% of the heavy atoms.
\end{itemize}

\begin{figure*}[htbp]
\begin{center}
\includegraphics[width=0.65\textwidth]{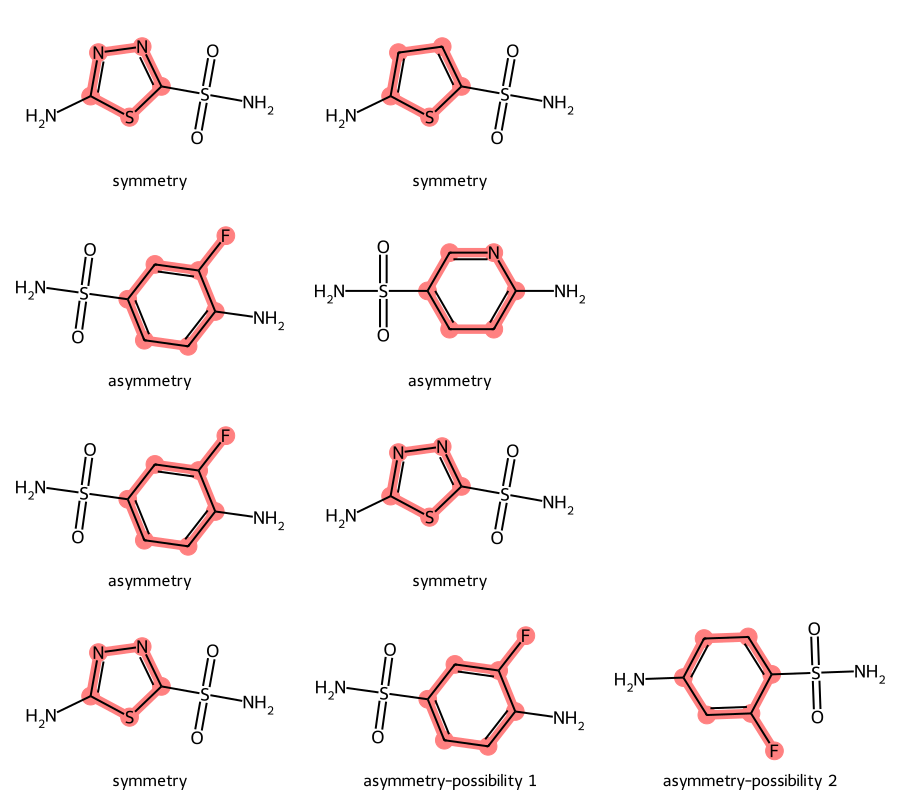}
\end{center}
\caption{Multiple possible permutations in symmetry-to-asymmetry core replacement.}
\label{fig:symmetry-to-asymmetry}
\end{figure*}

In cases where the original fragment exhibits symmetry and the replacement fragment does not (as illustrated in the last line of Figure~\ref{fig:symmetry-to-asymmetry}), multiple distinct products may arise due to different permutations of attachment point mappings, accounting for 7.5\% of our test set. Since the original attachment sites are chemically interchangeable, all such permutations are treated as valid ground truths for evaluation.

\paragraph{2. terminal replacement.}  
MMPs were classified as single-point replacements (or terminal replacements) if the number of heavy atoms in the modified fragment accounts for no more than 30\% of the initial molecule and only one attachment point.

\subsection{t-SNE visualization of fragment chemical space} \label{app:tsne-fragment-diversity}

\begin{figure*}[htbp]
\begin{center}
\includegraphics[width=0.8\textwidth]{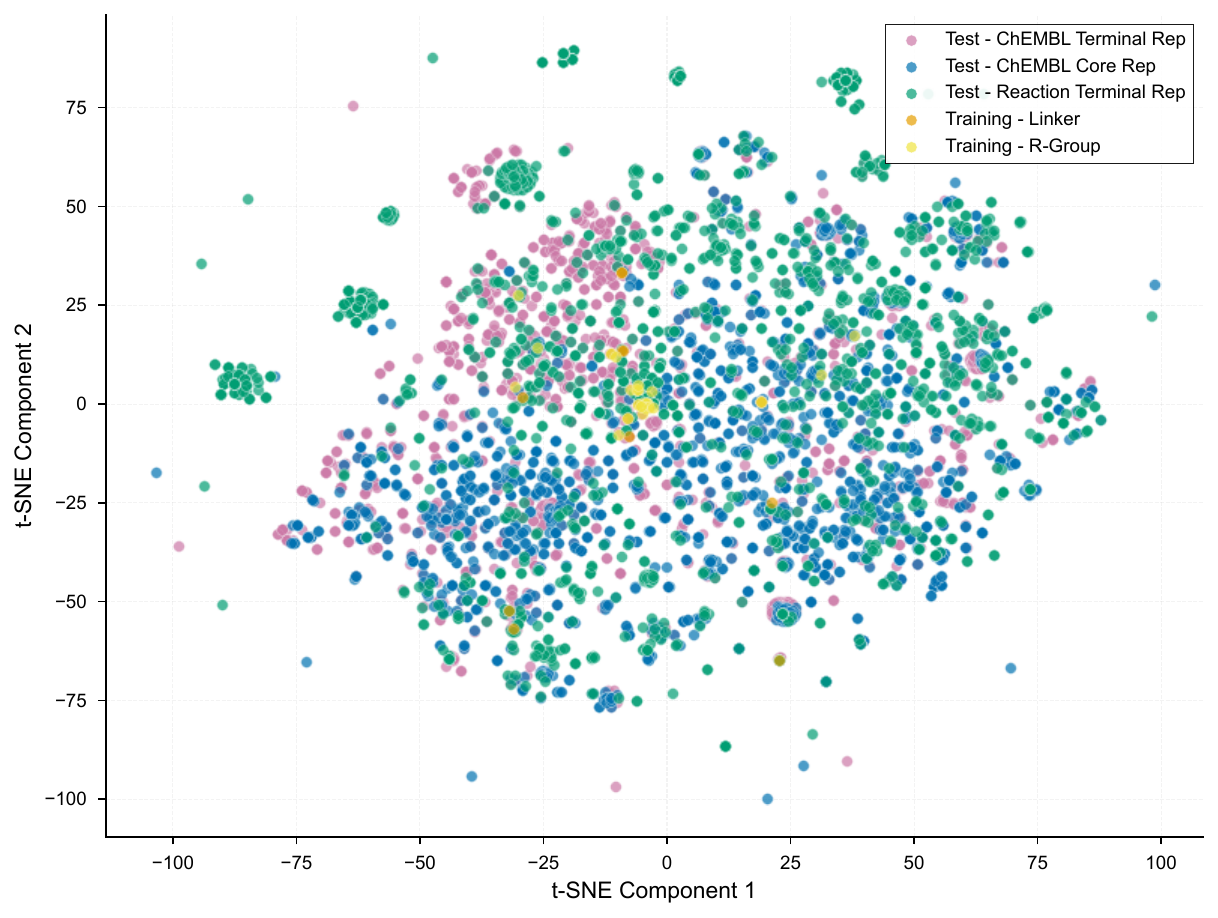}
\end{center}
\caption{t-SNE for fragments in synthetic training data construction and realistic edit test set.}
\label{fig:frag-tsne}
\end{figure*}

To assess the chemical diversity of the Realistic Edits test set relative to the synthetic moiety pool used in training, we computed ECFP fingerprints (radius = 2, 2048 bits) for all fragments and projected them into two dimensions using t-SNE. As shown in Figure~\ref{fig:frag-tsne}, fragments from the synthetic training pool cover only a small portion of the chemical space and form several compact clusters, reflecting the limited structural motifs used to construct the synthetic edit pairs. In contrast, fragments from the Realistic Edits set are widely dispersed across the projection space, occupying diverse regions that are largely non-overlapping with the synthetic clusters.

This spread confirms that Realistic Edits cover a substantially broader and more diverse chemical space. Despite this distribution shift, the finetuned code LLM in \abbr{} maintains strong performance, supporting our claim that the model learns generalizable code transformation patterns rather than relying on fragment-level memorization.

\section{Model training details} \label{app:model-training}

\subsection{Prompt format and design rationale} \label{app:prompt-prefiltering}

To determine an optimal prompting strategy, we conducted preliminary experiments examining the effects of various formatting choices. These preliminary experiments were conducted under conditions that differ from our final evaluation tasks; they serve only as pilot studies to guide the choice of attachment-point style.

For numbering attachment points in the SMILES, we tested:
\begin{itemize}
    \item Mark attachment points only in source SMILES with atom map numbers;
    \item Numerically number all atoms in SMILES and give the corresponding numbers of attachment points.
\end{itemize}
For marking attachment points in the fragment, we wrote original and target fragments with:
\begin{itemize}
    \item atom map number, e.g., [C:1]N
    \item rooted at attachment point with a short dash, e.g., -CN
    \item an asterisk mark, or a so-called dummy atom, e.g., *CN
\end{itemize}

\begin{table*}[ht]
\centering
\begin{tabular}{l l l c c}
\toprule
\multirow{2}{*}{\textbf{Action/Model}} & \multicolumn{2}{l}{\textbf{Molecule Number Style}} & \textbf{Attachment Point} &
\textbf{Whole SMILES}\\
\cmidrule(lr){2-3}
& \multicolumn{2}{l}{\textbf{Fragment Attachment Style}} & & \\
\midrule

\multirow{3}{*}{CodeGen / Coder}
    & Numbered & {[C:1]N} & 0\% & 0\% \\
    & Rooted  &  -CN & 20\% & 1\% \\
    & Asterisked  &  *CN & 6\% & 6\% \\
\midrule

\multirow{3}{*}{Direct / General}
    & Numbered & {[C:1]N} & 0\% & 3\% \\
    & Rooted  &  -CN & 12\% & 6\% \\
    & Asterisked  &  *CN & 4\% & 4\% \\
\bottomrule
\end{tabular}
\caption{Accuracy across different SMILES attachment styles and numbering strategies.}
\label{tab:pretest}
\end{table*}

Preliminary results are shown in Table~\ref{tab:pretest}. While the asterisk symbol (\texttt{*}) was not always the most accurate in pilot trials, it was retained because of its conventional role in representing dummy atoms in cheminformatics toolkits (e.g., RDKit, ChemDraw) and its compatibility with code generation. We also observed that passing numbered SMILES directly to the reasoning LLM leads to more stable responses, likely due to the limited ability of LLMs to handle SMILES during reasoning. In contrast, the rooted attachment-point style is not compatible with fragments requiring multiple attachment points, such as in core-replacement tasks. Our final prompt for code-based molecular editing is listed in Table~\ref{tab:edit_prompt}.

Generating both natural language rationales and SMARTS annotations introduces an additional structured representation requirement compared with using only natural-language editing intentions. However, this design does not simply transfer the difficulty of full-molecule SMILES generation to the intention stage. Instead, MECo decomposes molecular modification into short, local, and indexed edit specifications followed by explicit graph-level execution.

Pure natural-language intentions may be easier to express, but they are not always directly executable without ambiguity. For example, "replace the hydroxyl group" is underspecified when multiple similar groups are present, while symmetric and multi-attachment transformations, such as replacing a heteroaromatic core, require explicit correspondence between attachment sites. SMARTS-style fragments and atom indices are therefore introduced to reduce execution ambiguity and provide direct compatibility with RDKit. A natural-language-only formulation would require an additional grounding step from free-text descriptions or chemical names to molecular substructures before execution, potentially introducing errors in name resolution, functional-group localization, and attachment mapping.



% ...






\begin{table*}[htbp]
\begin{tcolorbox}[notitle, sharp corners, colframe=Periwinkle, colback=white, 
       boxrule=3pt, boxsep=0.5pt, enhanced, 
       shadow={3pt}{-3pt}{0pt}{opacity=1,mygrey},
       ]
\begin{verbatim}
You are given a molecule in SMILES format: 
"{numerically_numbered_source_smiles}". 

For reference, atoms where new groups will be attached are marked with 
"[*:n]", where "n" is the atom mapping number. 

Future connected atoms in groups are labeled using the same numbers, 
ensuring one-to-one attachment correspondence. You will then be given 
multiple instructions on how to edit the molecule.

Replace the substructure corresponding to 
"{original_fragment_smarts_1}" 
connected at atom {numbers_indicating_attachment_points} 
with "{target_fragment_smarts_1}".

...

Replace the substructure corresponding to 
"{original_fragment_smarts_n}" 
connected at atom {numbers_indicating_attachment_points} 
with "{target_fragment_smarts_n}".

Generate a Python code snippet that performs these replacements 
using RDKit via ChemicalReaction.

Ensure the code is executable and returns the modified molecule as
a new SMILES string.

You must return a Python code snippet wrapped in triple backticks.

The code should import modules from RDKit, perform the operation, 
and print only the modified molecule in SMILES format.
\end{verbatim}
\end{tcolorbox}
\caption{Final prompt for code-based molecular editing}
\label{tab:edit_prompt}
\end{table*}




\subsection{Training setup}

We used the official fine-tuning framework from the Qwen2.5-Coder repository%
\footnote{\url{https://github.com/QwenLM/Qwen2.5-Coder/tree/main/finetuning/sft}} and conducted full-parameter supervised fine-tuning (SFT) using DeepSpeed on both \texttt{Qwen2.5-Coder-7B-Instruct} and \texttt{Qwen2.5-7B-Instruct} models. The training was performed on 4 NVIDIA A100 80GB GPUs, with each epoch taking approximately 4 hours.

\section{Experiment details}

\subsection{Execution accuracy across fragment similarity bins} \label{app:acc_across_sim}

\begin{figure*}[htbp]
\begin{center}
\includegraphics[width=1.0\textwidth]{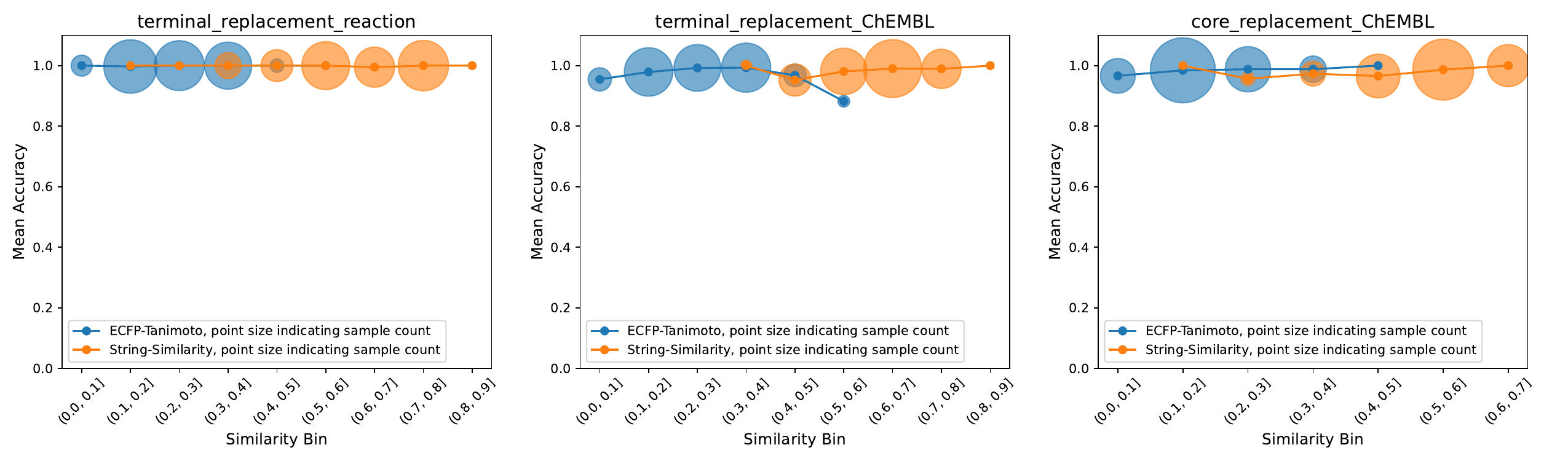}
\end{center}
\caption{Execution accuracy of the finetuned code LLM (Qwen2.5-Coder-7B-Instruct-FT) across bins of structural similarity (ECFP–Tanimoto) and string similarity (SequenceMatcher), evaluated on data from different sources and for both terminal and core replacements. The size of the semi-transparent halo around each point reflects the number of test samples in that bin.}
\label{fig:acc_sim_bins}
\end{figure*}

To further assess whether \abbr{} generalizes beyond the structural motifs present in its synthetic training set, we evaluate execution accuracy across fragment similarity bins computed using both structural similarity (ECFP–Tanimoto) and string similarity (SequenceMatcher ).
Figure~\ref{fig:acc_sim_bins} reports accuracy curves over a wide range of similarity scores, covering multiple realistic edit test sources, including ChEMBL for activity-derived edits, USPTO for reaction-derived edits, and both terminal and core replacement types.

The results show that \abbr{} maintains stable performance across the entire similarity spectrum, including bins where fragments are substantially dissimilar from those in the synthetic moiety pool.
This provides further evidence that \abbr{}’s execution ability is tied to neither structural coverage nor string pattern overlap of fragments in the training data, but instead generalizes robustly to structurally diverse and previously unseen edits.

\subsection{Implementation details for the molecular modification task} \label{app:prom_wrap}

ChemCoTBench~\citep{li_ChemCoTBench_2025} comprises six molecular modification tasks: three involving physicochemical properties (penalized logP~\citep{gomez2018automatic}, solubility~\citep{delaney_esol_2004}, QED~\citep{bickerton_qed_2012}, implemented in RDKit~\citep{landrum_rdkit_2013}) and three involving target-specific bioactivities (DRD2~\citep{olivecrona_drd2_2017}, JNK3, and GSK3$\beta$~\citep{li_jnk-gsk_2018}, implemented in TDC Oracles~\citep{huang_tdc_2021}). Each task provides a set of input molecules ${M_i}$ paired with a shared optimization goal $T$. This setting reflects real-world design scenarios, where high-level property goals must be translated into concrete molecular modifications. For each $(M_i, T)$ pair, we compare direct molecule generation from a reasoning LLM with the MECo framework, which first generates an editing intention and then executes it via code. This evaluation allows us to assess whether MECo improves over direct generation approaches in terms of (1) chemical validity; (2) optimization success rate; (3) mean similarity to source molecules; and (4) structure–action consistency, which is computed only for human-interpretable cases.

A prompt wrapper is used to instruct LLMs to generate modification schemes together with the underlying rationales, rather than merely producing SMILES strings, as shown in Table~\ref{tab:prompt_wrapper}.


\begin{table*}[htbp]
\begin{tcolorbox}[notitle, sharp corners, colframe=Periwinkle, colback=white, 
       boxrule=3pt, boxsep=0.5pt, enhanced, 
       shadow={3pt}{-3pt}{0pt}{opacity=1,mygrey}]
\begin{verbatim}
{Objective description}

Your response must be actions to perform the transformation, i.e. 
"replace original_group_smiles connected at atom_number with
target_group_smiles". 

Each action must include two SMILES strings indicating the original
group and the target group.

Use dummy atoms to mark the connection point, e.g. 
"replace [*:1]Cl connected at atom_number with [*:1]OC" 
or 
"replace [*:2]c1ccc([*:3])cc1 connected at atom_number2, atom_number3 
 with [*:2]c1cnc([*:3])cc1".

Your response must be in directly parsable JSON format:
{
    "Action Description": [
        "replace original_group_smiles connected
        
        at atom_number with target_group_smiles",
       
        ...
    ],
    "Final Target Molecule": "SMILES"
}

Given the source molecule with atoms numbered:
{number_smi(source_smiles)}.

You should ignore Hydrogen (H) and numbers in the group SMILES.
\end{verbatim}
\end{tcolorbox}
\caption{Prompt wrapper in the molecular modification task.}
\label{tab:prompt_wrapper}
\end{table*}

All dataset sampling, model training, and local inference used a fixed random seed of 42. For API-based models, we used a fixed seed of 42 and greedy decoding where supported. GPT-5 and GPT-4o were accessed through Microsoft Azure OpenAI using API versions \texttt{2025-01-01-preview} and \texttt{2024-08-01-preview}, respectively. DeepSeek-R1-250528 and deepseek-v3-241226 were accessed through Volcengine Ark, and Gemini-2.5-Pro and Gemini-2.0-flash through Google Vertex AI. All unspecified API parameters were left at their service defaults.

To quantify evaluation uncertainty, we performed paired bootstrap resampling over the existing instance-level outputs. Macro-average improvements of \abbr{} over direct generation from DeepSeek-R1 are 31.2 points in validity with 95\% CI [26.7, 35.5], 24.2 in success [19.8, 28.5], 0.256 in similarity [0.231, 0.281], and 70.2 points in CR [66.2, 74.1]; all remain significant after Holm correction.

\subsection{Manual evaluation of consistency} \label{app:manual_criteria} 

\paragraph{Criteria.} We considered syntax errors in actions that are still discernible to human experts. Examples include malformed atom map annotations such as \texttt{[*1:]} or \texttt{[*1]}, which should be \texttt{[*:1]}, and missing explicit hydrogens on aromatic nitrogens, e.g., \texttt{[*:1]c1nnnn1}, which should be \texttt{[*:1]c1[nH]nnn1}. 

For attachment point checks, we adopted relatively loose criteria. For instance, in \texttt{[cH:1]1[cH:2][cH:3][cH:4][cH:5][c:6]1}
\texttt{[O:7][CH2:8][CH:9]1[CH2:10][CH2:11]1}, assigning the attachment of \texttt{[*:1]OCC1CC1} to either atom 6 or 7 is considered discernible. Similarly, in \texttt{[*:1]O[*:2]}, providing only one side of the atom index (e.g., 6 or 8) or the atom index alone (7) is also regarded as discernible. By contrast, edits that invoke groups not present in the structure are considered invalid, even if similar patterns can be found.

\begin{figure*}[htbp]
\begin{center}
\includegraphics[width=0.8\textwidth]{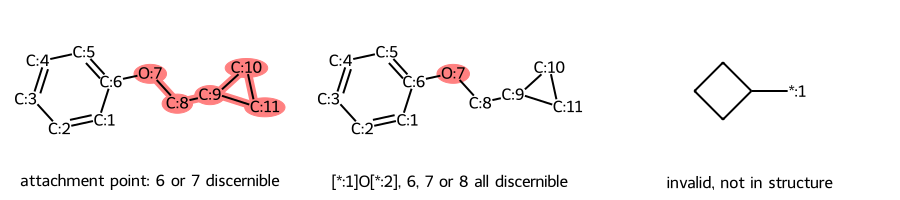}
\end{center}
\caption{Visualization of manual check criteria.}
\end{figure*}

Nevertheless, manual checking is inherently subjective. Different researchers may interpret borderline cases differently, and there remains a possibility of human error.

\paragraph{Subjectivity.}
To further assess potential subjectivity, we adopt a blinded multi-annotator protocol: three annotators independently evaluate 120 samples (evenly drawn from DeepSeek-R1), without access to model identities, task provenance, or one another’s labels. Labels were likewise highly consistent: three-way agreement = 95.83\%, two-way = 3.33\%, and complete disagreement = 0.83\%. Attachment-point partiality judgments achieved Fleiss' $\kappa$ = 0.9342 for the 104 jointly valid cases, indicating near-perfect agreement.

\paragraph{Scalability.} This verification pipeline naturally supports scalable dataset growth: for each natural-language edit instruction, annotators provide one gold-standard structure (e.g., SMILES) that correctly implements the edit. Each labeled (instruction, targeted structure) pair becomes a reusable test case and can be automatically applied to any future code-LLM output via structure matching. Thus, each round of human labeling incrementally expands a standardized, reproducible, and low-cost evaluation suite. In addition, the annotation process can be substantially accelerated through formalized validation code. For example, substructure errors (e.g., fragments not present in the source molecule) can be automatically excluded before human inspection. This further reduces subjective load and ensures that annotators only review chemically valid candidate edits.

Regarding scalability toward larger and more diverse test sets, we draw on recent progress in LLM evaluation and outline several feasible extensions:

1. \textbf{LLM-as-a-Judge with minimal supervision.}
   \citet{lin_llm-eval_2023,li_hypoeval_2025} show that structured prompting enables LLMs to reliably approximate human judgments at scale, suggesting that \abbr{}'s intention-consistency checks could be extended through supervised automatic judges.

2. \textbf{Decomposed, verifiable evaluation dimensions.}
   \citet{zhou_instruction-following_2023,ye_flask_2023} show that breaking complex behaviors into verifiable atomic criteria yields more stable and interpretable assessments. Future \abbr{} versions may evaluate edits along dimensions such as (i) substructure localization, (ii) bond/valence correctness, (iii) atom-mapping agreement, and (iv) chemically valid topology, paired with automatic graph/SMILES validators. This reduces subjectivity and aligns evaluation with chemically interpretable components.

3. \textbf{Standardized benchmark construction.}
   Successful evaluation suites~\citep{zhou_instruction-following_2023,zheng_llm-as-a-judge_2023} emphasize fixed task sets, unified templates, and executable scoring pipelines. Inspired by these designs, we could build a standardized benchmark for molecular edit–intention consistency with public task suites and unified evaluation scripts, ensuring fully reproducible cross-model comparison.

These extensions provide a clear and technically grounded path toward scaling \abbr{}'s intention-consistency evaluation beyond the manually assessed subset while enhancing reproducibility, transparency, and objectivity.

\subsection{Observation on terminal replacement test samples} \label{app:ob_ter_re}

In Section~\ref{subsec:edit_real}, we noted a discrepancy between reaction-derived and target-specific terminal replacements. To investigate this further, we conducted a closer inspection and found that the discrepancy may stem from differences in SMILES syntax patterns: in target-specific edits, dummy atoms (e.g., \texttt{[*:1]}) often appear in the middle or at the end of SMILES strings, whereas in reaction-derived edits, they frequently occur at the beginning. This difference arises from the implementation of the extraction program rather than the underlying data distribution, meaning that the two syntax types can be transformed into each other by choosing whether the dummy atom or another atom is used as the SMILES root. Such pattern sensitivity suggests that LLMs struggle to generalize over SMILES syntax, further motivating the use of structured code as a more robust and interpretable intermediate representation.

\subsection{Qualitative Error Analysis}

Qualitative inspection of failed examples reveals representative failure modes on both the code-generation and reasoning sides. On the code-generation side, as discussed in Section~\ref{app:ob_ter_re}, performance is sensitive to dummy-atom placement, particularly when attachment markers appear at the end of complex SMARTS expressions.

On the reasoning side, failures in the downstream end-to-end setting can arise from structured editing intentions that are difficult to execute reliably. Occasional atom-index offsets may either cause editing to fail or mislocalize an otherwise valid editing intention, with the error subsequently propagating to execution. Other cases lie at the boundary between reasoning and execution. For example, the reasoner may propose atom-level substitutions within aromatic systems, where lowercase SMARTS atoms encode aromaticity. Such intentions can be readily understood by a human expert, but their realization through reaction-based editing may require reformulating the edit into a reaction-compatible operation. Since the current lightweight coder is designed for faithful execution rather than semantic reinterpretation and reimplementation, these cases remain challenging.

\subsection{Claude-Sonnet-5 on Realistic Edit Benchmarks}

To assess whether the code-based editing formulation can be effectively executed by a leading frontier code LLM without task-specific fine-tuning, we additionally evaluated Claude-Sonnet-5 on 100 randomly sampled examples from each realistic edit category. It achieves accuracies of 98.0\%, 62.0\%, and 73.0\% on reaction-derived terminal replacement, target-specific terminal replacement, and target-specific core replacement, respectively.

\subsection{Property improvement vs.~similarity} \label{app:imp_sim}

\begin{figure}[!ht]
\centering
\includegraphics[width=\columnwidth]{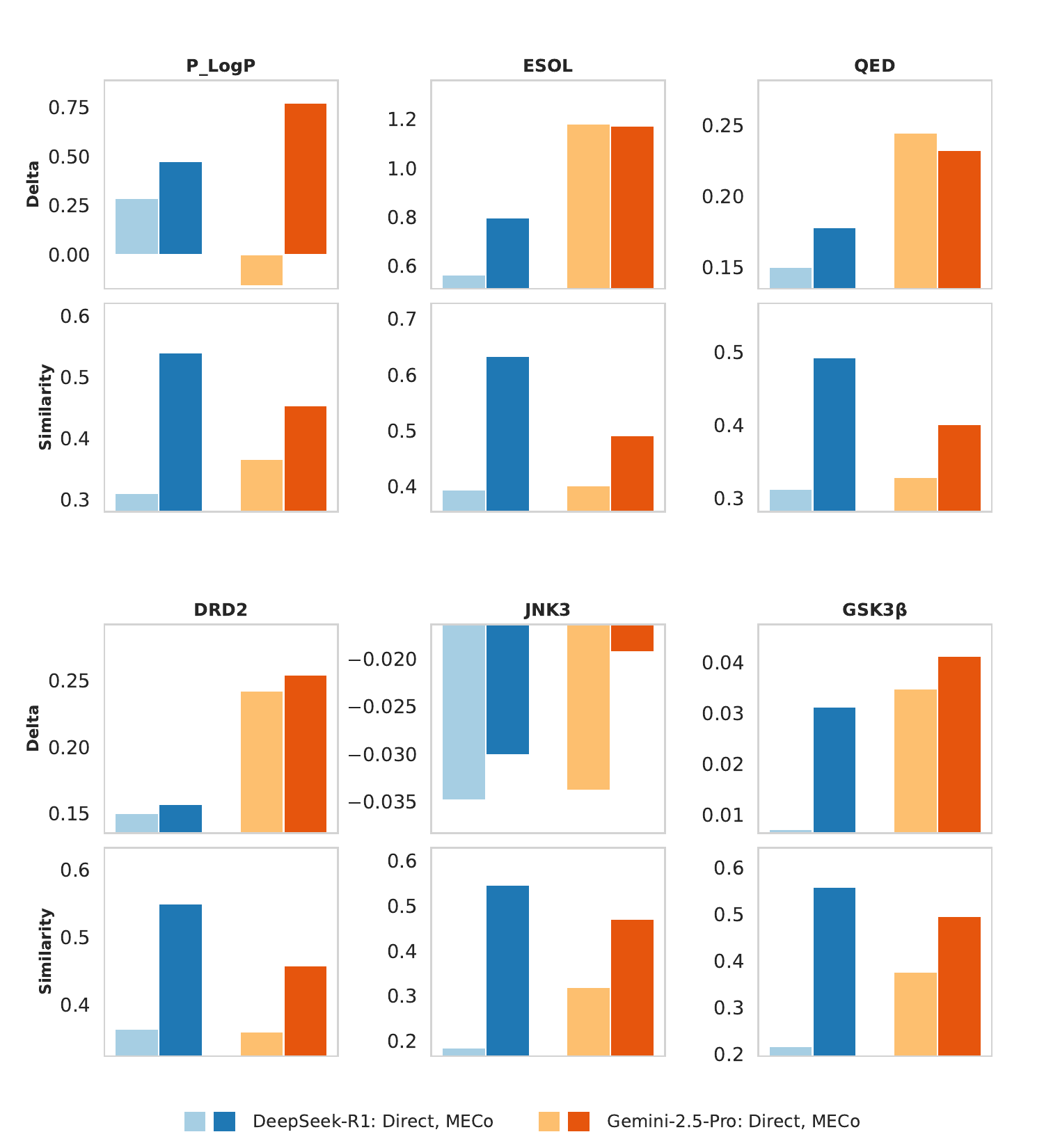}
\caption{Bar plots of molecular optimization performance across six tasks, showing property improvement (top row) and structural similarity (bottom row) for Direct generation and \abbr{} for DeepSeek-R1 and Gemini-2.5-Pro.}
\label{fig:imp_sim}
\end{figure}

To further analyze the trade-off between property improvement and structural preservation, we compare the results of direct generation and \abbr{} separately on all six tasks (Figure~\ref{fig:imp_sim}). The top row shows the average property improvement (relative to the initial molecule), where \abbr{} generally achieves higher improvements than direct generation. The bottom row shows the average structural similarity (ECFP4-based) to the initial molecule. While Gemini-2.5-Pro exhibits larger improvements in the upper row, it tends to reduce similarity more than DeepSeek-R1, indicating that different LLMs show different levels of willingness to sacrifice structural similarity in exchange for property gains. This observation motivates our use of success rate as the primary evaluation metric in the main paper, as it balances both optimization effectiveness and structural plausibility.

\subsection{Docking setups for case study}

For the case study, molecular docking was performed using the Glide SP (standard precision) protocol implemented in the Schrödinger suite. The high-resolution X-ray co-crystallized target protein structure was obtained from the Protein Data Bank (PDB ID: 4AFJ).  

Protein preparation was carried out using the Protein Preparation Wizard, including addition of hydrogen atoms, assignment of bond orders, optimization of the hydrogen-bonding network, and restrained minimization of heavy atoms with the OPLS\_2005 force field. Ligands were prepared using LigPrep to generate at most 32 stereoisomers.

Docking was performed with default Glide SP parameters unless otherwise specified. The receptor grid was centered on the co-crystallized ligand SJJ in 4AFJ. The top-ranked docking poses were retained for analysis and visualization.

\end{document}